%% file: samplepaper.tex
\documentclass[runningheads]{llncs}
\usepackage[T1]{fontenc}
\usepackage{graphicx}
\usepackage{subfiles}

\begin{document}
\title{Transformers As Approximations of Solomonoff Induction}
\author{Nathan Young\inst{1}\orcidID{0000-0001-6695-6441} \and
Michael Witbrock\inst{1}\orcidID{0000-0002-7554-0971}
% \and Robert Amor\inst{1}\orcidID{0000-0002-4329-9044}
}
\authorrunning{N. Young, M. Witbrock}
% \authorrunning{N. Young et al.}
\institute{University of Auckland, Auckland, New Zealand\\
\email{\{nathan.young,m.witbrock\}@auckland.ac.nz}}
\maketitle              % typeset the header of the contribution
\begin{abstract}
Solomonoff Induction is an optimal-in-the-limit unbounded algorithm for sequence prediction, representing a Bayesian mixture of every computable probability distribution and performing close to optimally in predicting any computable sequence.

Being an optimal form of computational sequence prediction, it seems plausible that it may be used as a model against which other methods of sequence prediction might be compared.
    
We put forth and explore the hypothesis that Transformer models - the basis of Large Language Models - approximate Solomonoff Induction better than any other extant sequence prediction method.
We explore evidence for and against this hypothesis, give alternate hypotheses that take this evidence into account, and outline next steps for modelling Transformers and other kinds of AI in this way.

\keywords{Explainability  \and Interpretability \and Solomonoff Induction.}
\end{abstract}

\subfile{0_introduction}

\subfile{1_background}

\subfile{2_hypothesis}

\subfile{3_for}

\subfile{4_against}

\subfile{5_synthesis}

\subfile{6_next}

\subfile{bib}
\end{document}

%% file: 0_introduction.tex
\section{Introduction}
It has become something of a cliché for papers in the field of Artificial Intelligence to begin by recognising the rapid progress the field has made in the past few years; progress made possible by the ubiquitous Transformer architecture. Transformers - the basis of Large Language Models (LLMs) - seem to scale without limit, achieving greater performance with greater size and greater amounts of training data.

However, despite their power and ubiquity, Transformers are still poorly understood. Many papers have been written and experiments carried out in pursuit of making these models interpretable, understandable, and explainable, with some success - in particular, identification of patterns identified by particular neurons~\cite{goldengate}~\cite{LLMsOnLLMs}. But such work is largely ad hoc, with no common approach. The field of interpretability, as well as AI safety research in general, remains preparadigmatic.

Having a standardised way of analysing the performance or structure of black box algorithms like Transformers requires a standard model to compare against - some generalisation of Transformers and other sequence predictors, of which all other models can be considered departures.
One class of such models are \textit{unbounded solutions}; that is, solutions to a given problem that ignore computational constraints.

One influential example of such an algorithm is Shannon's 1950 solution to computer chess~\cite{shannon}: Shannon's model evaluates the entire game tree, evaluating which moves lead to an optimal outcome if both players play optimally.
This requires an absurd amount of memory, but modern chess algorithms evaluate smaller sections of the game tree and use other heuristics to estimate the value of the resulting board layouts.
This illustrates how unbounded algorithms can be invaluable theoretical ideals, even when they are useless in practice.

Such an ideal unbounded model of sequence prediction exists, and was proposed in 1964 by Ray Solomonoff.~\cite{solind} Solomonoff often referred to his conception as "algorithmic probability", but it is more commonly known today as Solomonoff Induction (which we abbreviate as SolInd).
In contrast to Shannon's chess algorithm, which was merely impractical, SolInd is incomputable without infinite memory and the capability to execute an infinite number of programs in a finite time period.
SolInd is considered to be an optimal computable sequence prediction scheme, making only a finite amount of unforced prediction error on any computable sequence relative to any other prediction scheme.

In this paper, we propose that Transformers - and, indeed, all sequence prediction schemes - should not just be measured against SolInd, but modelled and understood as direct approximations thereof.
Of course, actual implementations of SolInd - using infinite computing power and infinite memory - are incomputable and impossible within a finite observable universe, and so no practical algorithm can ever implement it directly.
Still, we propose that Transformers and Neural Networks (NNs) in general can be thought of as bounded approximations of SolInd, with a similar basic underlying structure supplemented by optimisations not present in the unbounded model.

We will outline some specific reasons to believe that Transformers approximate SolInd, as well as some findings that appear to contradict this hypothesis. We conclude by synthesising these results and outlining next steps for exploring this way of modelling Transformers.

%% file: 1_background.tex
\section{Background}
We will begin by outlining, in short, how Solomonoff Induction works.

We first consider a Universal Turing Machine, or UTM, which we call $M$, with a unidirectional input tape, a bidirectional working tape, and a unidirectional output tape. For simplicity, we will assume that $M$ uses a binary alphabet, but these results generalise to any alphabet.
Any input string $s$ given to $M$ will result in some output string $x$. Since $M$ is capable of simulating any Turing Machine (TM), there exists an $s$ that will generate any arbitrary $x$. The a priori probability of $s$ being given as input to $M$ is considered to be the length $l(s)$ of $s$ in bits.
Since $M$ is likely to give $x$ on many different inputs, we consider the set of minimal input strings $S$ that produce $x$ when given to $M$ (possibly followed by other characters).

The probability $P$ (with respect to machine $M$) of a string $x$ is given as:

$$P_M(x) = \sum_{s\in S} 2^{-l(s)}$$

The smallest such $l(s)$ will usually make up the majority of this sum, so considering only the smallest description of a given program gives a reasonable approximation.

The probability of a given continuation $y$ of $x$ - in other words, the probability of the output string beginning with $xy$, given that $M$ has only produced $x$ so far - is simply:

$$P_M(xy|x) = \frac{P_M(xy)}{P_M(x)}$$

In this way, Solomonoff Induction represents a Bayesian prior probability on every computable string, with each new symbol in the input string triggering a Bayesian update.

An equivalent form of SolInd, given in \cite{solind}, considers Probability Evaluation Methods (PEMs) instead of TMs, updating probability weights on particular inputs rather than eliminating those that do not produce $x$.
This conception of SolInd gives equivalent results to the version given above.

SolInd's speed of converging on any given sequence depends on the choice of UTM - if a sequence $x$ has a shorter description on $M$ than on some other UTM $N$, then $M$ will make less prediction error on $x$ than $N$ will. This has limited importance in the limit, though it may be a vital consideration in practice.

SolInd is considered optimal in the following ways:
\begin{itemize}
    \item Bounded error: the amount of prediction error SolInd will make on any given sequence (measured in log odds to base 2, making errors comparable with bits) is guaranteed to be less than $P(x)$.
    \item Universality: SolInd's predictions will converge to optimality on every computable string.
    \item Pareto optimality: while other prediction schemes can predict some sequences better than SolInd, it is impossible for any method to predict a sequence better than any SolInd without predicting some other sequence worse.
\end{itemize}

Proofs of these claims and further analysis of SolInd's performance can be found in \cite{newbounds} and \cite{solind96}.

%% file: 2_hypothesis.tex
\section{Hypothesis}
We now outline our key hypotheses, beginning with our reasoning and ending with their significance to AI research.

\subsection{Reasoning}
The basis of this paper was perhaps best expressed by Solomonoff himself in a paper published 32 years after his initial description of SolInd:\cite{solind96}
\begin{quote}
    Just about all methods of induction can be regarded as approximations to [SolInd], and from this point of view, we can criticize and optimize them. Criticism is always easier if you have an idealized (possibly unrealizable) standard for comparison. [SolInd] provides such a standard.
\end{quote}

In other words, it makes sense to analyse any sequence prediction method by comparing it to SolInd, considering any architecture not found in SolInd to be an optimisation that helps it achieve performance closer to SolInd's theoretical ideal.

Our main consideration is in comparing Transformer models to SolInd, since they currently represent an unopposed state-of-the-art in almost all fields of machine learning and artificial intelligence; however, the intention is to introduce a way of understanding every method by which continuations of sequences can be predicted, including all those invented to the present day and all those which might be invented in the future.
Indeed, it is necessary to compare Transformers' approximations of SolInd to other prediction methods if we are to understand how they achieve their superior performance.
% So it might make sense to think of practical sequence predictors as being closer to or further from it
% In fact this may be a reasonable standard to compare literally every sequence predictor to

\subsection{Statement of hypotheses}

We now outline our main hypotheses that we will interrogate in this paper and in future work:

\begin{enumerate}
    \item Transformers can be modeled as approximations of Solomonoff Induction.
    \item Transformers approximate Solomonoff Induction more closely than other methods of sequence prediction (e.g. other neural networks).
    \item Transformers can be modeled as approximations of Solomonoff Induction at test time specifically.
    \item Stochastic gradient descent approximates Solomonoff Induction during training time, and does this better than other training schemes.
\end{enumerate}

The final hypothesis recognises a key difference between SolInd and machine learning systems; Solomonoff inductors are not 'trained' in the same way as NNs, and therefore must be thought of in a slightly different way. For example, when comparing SolInd and an NN, we might consider the SolInd input string $s$ to contain the entire training corpus used to train the NN. Therefore, we must consider how a NN's training approximates SolInd, as well as its operation at runtime.

\subsection{Significance}
We expect confirmation and/or refutation of our hypotheses to be consequential for the way Transformers and other machine learning systems are understood. Some possible implications of our hypotheses include:

\begin{itemize}
    \item \textbf{Improvements to architecture:} if the Transformer architecture is found to resemble SolInd in some important way that other NNs lack, this would explain their superior performance and possibly prescribe further improvements.
    \item \textbf{Explainability:} identifying specific ways that Transformers implement different TMs/PEMs and update weights on each would go some way towards explaining how Transformers work and why a particular output was given.
    (This point is predicated on the assumption that SolInd is naturally more easily explainable than a Transformer. We believe this is reasonable, but that identifying the purpose of a particular TM (and thus the output of any SolInd) is still non-trivial.)
    \item \textbf{Producing explicit models:} once key parts of a Transformer have been identified, it may be useful for both performance and explainability for these parts to be split into separate programs - no longer hiding in inscrutable matrices but operating more like traditional algorithms.
    \item Finally, if our hypothesis is refuted and Solomonoff Induction is shown to not be a useful model to compare prediction schemes against, the following questions must be raised: Why not? What about Solomonoff's optimal sequence predictor is so unlike its computable counterparts? And is there another model - unbounded or otherwise - against which arbitrary prediction methods \textit{can} be compared? 
\end{itemize}

%Okay but so what if it is?
% Could have implications for improvments to architecture (eek! focus on how this could make them more explainable, rather than more effective)
% Could improve explainability
% Could lead to models being made that aren't inscrutable matrices but are more explicitly Solomonoffesque, splitting capabilities up into separate modules
% Could help optimise existing models??
% If this turns out to not be a helpful lens at all, why? Is SolInd just too weird and incomputable to serve as an ideal for computable systems? Can we instead compare them to computable 

%% file: 3_for.tex
\section{Findings in favour}
In this section, we will present several findings that support and shed light on our hypotheses.
% Remember that one about pruning that you talked to that guy at LessOnline about

\subsection{Universality}
One key result on which our hypothesis relies is the idea that Transformers - and NNs in general - can implement arbitrary programs (equivalently, can emulate arbitrary TMs).
That Recurrent Neural Networks (RNNs) can emulate arbitrary TMs was first shown in 1992 by Siegelmann and Sontag~\cite{siegelmannsontag}. This result has since been replicated for many other machine learning models; indeed, Transformers have had multiple papers showing distinct ways in which they might emulate arbitrary TMs.\cite{tftm1}\cite{tftm2}

Therefore, one way that Bayesian mixtures of TMs could be emulated using a NN is for them to be simply averaged; given a probability distribution over a set of TMs, construct emulations of each on a NN of a certain size and give each weight in the NN its average value among the emulations, weighted by that emulation's probability mass.
It is possible that the class of weights that can be achieved by weighted-averaging emulated TMs in this way is equivalent to the class of all randomised weights that can be given to a NN.

%And then there's the possibility of emulating multiple at once, ooo, neato

\subsection{Decomposition}
There exists a substantial amount of research into NN decomposition~\cite{decomp}; that is, taking a NN trained for some task (or set of tasks) and identifying smaller subnetworks that can perform the task without the rest of the network (or perform one of the set of tasks, with other parts of the network possibly performing other tasks).

One well-supported hypothesis in this space is the \textit{lottery ticket hypothesis}~\cite{lottery}: that randomly-initialised networks contain subnetworks that can be achieve similar performance to the whole network if trained in isolation, having been given random values that quickly reach a high level of performance.
This implies that such "winning tickets" exist as part of any random initialisation of any NN, for every function that that NN is capable of emulating.
Further, since a RNN can emulate a UTM with as few as 1058 neurons~\cite{siegelmannsontag} - many fewer than modern deep NNs - most functions can be reasonably considered to be emulable by most NNs.

In other words, there is good reason to believe that randomly-initialised networks contain something close to a representation of any arbitrary TM, the most relevant of which can be found during training and given increased weight over other subnetworks. This seems to closely mirror SolInd's process of induction.
% People have already tried splitting TFs into separate computational modules, with some success! Looks SolIndy to me
% Also shows the whole composition thing to be legit - can't decompose if you don't compose first
% Also, NNs trained to do a simple task can usually be pruned to a much smaller subset that can achieve equal performance - if trained on multiple, might be doable multiple times

%% file: 4_against.tex
\section{Findings against}
In this section, we will explore findings that seem to contradict or complicate our hypothesis.
Of course, computable instantiations of Solomonoff Induction are impossible, so the strongest version of our hypothesis, that Transformers act as Solomonoff Inductors, cannot be absolutely true; any attempt to approximate SolInd will be subject to limitations.
These findings can be thought of as exploring and naming these limitations.

\subsection{Limits of stochastic gradient descent}
Computational tasks can be sorted into four categories: regular, context-free, context-sensitive, and recursively enumerable. These categories each represent an increase in the level of complexity required to complete them, forming the Chomsky hierarchy.~\cite{chomsky}

After training on tasks at varying levels of the Chomsky hierarchy, Delétang et al.~\cite{deletang} found that Transformers can generalise well to regular tasks, but suffer greater performance penalties on tasks that are further up on the Chomsky hierarchy - eventually leading to recursively enumerable tasks, which correspond to those that can be solved by Turing machines. Other kinds of NNs tested exhibited similar behaviour.
The authors speculate that this may be due to the limitations of stochastic gradient descent as a training technique - when weights are gradually changed to approach an ideal value, small imperfections can accumulate.

% It can't actually get to every TM, if that matters

\subsection{Computational limits in practice}
NNs failing to generalise may also be explained by a fundamental limitation of the way they emulate TMs.

To emulate arbitrary Turing machines, an infinitely large tape is required. This can be achieved in NNs with infinite width, but is more commonly achieved by using the value of particular neurons as memory, with the decimal expansion of the neuron's value (in some base) forming a stack.
For this method to achieve arbitrary memory size requires unlimited precision, which is rarely achieved in practice. This may or may not be a relevant limitation in practice; it does not stop finitely large computers from running useful programs, but it may shrink the amount of memory available to a NN below what is needed for many programs.

Another fundamental limitation is that of state memory. If arbitrarily many NNs that emulate TMs are averaged to give a new NN that has the average weights of each (possibly weighted by their probability mass in a distribution, as outlined above), the resulting NN may be able to evaluate the composite transition function in finite time and space, but cannot track arbitrarily many state variables in finite space.
This may prevent the SolInd emulation method given above of averaging weights among emulated TMs in a given probability distribution, since their states will, without any extra work, all be tracked in the same set of neurons. This can be worked around if the TMs' states share meaningful information, but this is not a trivial requirement.

\subsection{Transformers make poor Solomonoff Inductors}
Finally, an attempt was made by Grau-Moya et al.~\cite{tfsolind} to train a Solomonoff Inductor using data sampled directly from a UTM. They trained multiple kinds of machine learning systems, including Transformers.
(This is an example of meta-learning; that is, training systems that learn new tasks quickly instead of simply training them on the tasks themselves. This is distinct from our work, which seeks to show that Transformers already have this capability to at least some degree without training.)

Among the models tested in this paper, Transformers often learnt best of all models tested when tested on sequences of lesser or equal length to that given to them at training time. However, when given longer sequences, their performance often became the worst of any model.\footnote{This may be explained by an architectural choice in this paper: Grau-Moya et al.'s Transformer model used Vaswani et al.'s original fixed positional encoding~\cite{tf}, rather than a relative encoding. This may have lead to the Transformer not being able to recognise patterns that were longer than 256 symbols long, as it had not previously seen positional encodings larger than 256. A relative positional encoding may have helped to address this problem.}
This suggests that Transformers may not be suited to learning and/or emulating arbitrary TMs in practice.

% It can't actually approximate every TM without infinite precision
% Also, like... have you actually tried to do this?? You can't just naively average the TM values, the memory will get super fucked up.
% Also, how about that DeepMind tried to do this and Transformers didn't work well, how about that
% Well we did find a concrete flaw in their model that would seem to explain it, so there's that

%% file: 5_synthesis.tex
\section{Syntheses and alternate hypotheses}
In this section, we will propose some alternate hypotheses, which are more complicated than those previously outlined but take the above findings into consideration.
% Okay so that's a bust but hey, there's no reason SolInd can't work with other things
% And maybe it's a l

\subsection{Alternate Solomonoff models}
We begin by exploring models that utilise the key insights of SolInd, while taking into account the fact that Transformers may be poor at (and possibly even incapable of) emulating arbitrary TMs.

SolInd represents a Bayesian mixture over all inputs to a UTM, and therefore all Turing machines. But there is no reason why we cannot consider similar probability distributions over alternative methods of computation, lower on the Chomsky hierarchy.

The idea that Transformers can more easily approximate simpler models than TMs, as discussed before in the context of Delétang et al.'s work~\cite{deletang}, was also explored by Liu et al.~\cite{shortcuts}, who found that Transformers are well-suited not only to learning functions that can be implemented in finite-state automata (i.e. regular languages), but to finding shortcuts to them that can yield identical solutions in logarithmic time compared to executing the automaton - that is, if the automaton would take $T$ steps to yield a solution, Transformers can learn to find the same solution in $O(T)$ time.

It may therefore make more sense to think of Transformers as a mixture of finite-state automata rather than TMs, or possibly other models of computation like Markov chains, pushdown automata, or bounded Turing machines.Since Turing machines are capable of simulating every possible form of computation, including these, it may still make sense to consider decomposing Transformers into TMs, if not \textit{all} TMs.
% Like Markov chains? Or simpler automata?
% Or maybe it's making some big practical optimisations, like picking just a few TMs??? Is that still a useful hypothesis??
% Ooh, can you make a Solomonoff model with finite working memory?? Probably not?? Not with TMs anyway, but you can with other models probably, especially with the whole shortcut thing *taps forehead*

% \subsection{Training as UTM selection}
% We 
% Where's training as UTM selection? Is it needed? It seems hard to fit in with everything else.
% Leave it off for now, write the conclusion, come back

\subsection{Alternate hypotheses}
In light of these considerations, we add the following to our list of hypotheses:
\begin{enumerate}
    \setcounter{enumi}{4}
    \item Transformers can be modelled as Solomonoff inductors with limited memory, preventing them from emulating all TMs and making it harder for them to solve tasks higher on the Chomsky hierarchy.
    \item Transformers can be modelled as Bayesian mixtures of simpler computational schemes than the class of all TMs.
    % \item UTM selection, while unimportant in theory, is vitally important in practice - perhaps even more so than updating during runtime.
    \item The further down a model of computation is on the Chomsky, the more quickly it is converged upon by stochastic gradient descent, leading Transformers to give much more weight to some classes of TM than others.
\end{enumerate}

It should be noted that these hypotheses are not all mutually exclusive, and that some middle ground may well explain Transformers better than any individual hypothesis alone.
For example, Transformers might represent a probability distribution over all finite-state automata, as well as executing some TMs that evaluate recursively-enumerable functions that simpler automata cannot approximate.

Still, it is useful to consider which hypotheses have more explanatory power - are Transformers more like TM-infused departures from mixtures of finite-state automata, or UTMs that encode automata well but can still approximate any TM with enough time and space? These two composite hypotheses would prescribe different implications towards explaining and decomposing Transformers, and so must be separated and carefully considered.
% Okay what about simpler automata?
% What about SDG as UTM selection?

% \subsection{Hypotheses not necessarily mutually exclusive}
% Of course it could be thought of as executing TMs in conjunction with other stuff, and it almost certainly does, this shit's messy af
% But maybe it's best to think of it as a TM-infused departure from a mixture of Markov chains

% This should maybe not be its own section. Paragraph or subsubsection, maybe?

%% file: 6_next.tex
\section{Conclusion}
In this paper, we have proposed that Transformers and other NNs should be analysed under the lens of Solomonoff's optimal unbounded sequence prediction scheme, Solomonoff Induction. Since SolInd represents the upper bound of what any sequence predictor can be capable of, it likely has extensive prescriptive and descriptive power with respect to NNs, but little work has been done to utilise this.

We have outlined reasons why this way of modelling NNs may be helpful in understanding how they work and achieving similar performance using algorithms that are less inscrutable and easier for programmers to work with. We have given several hypotheses as to how Transformers may approximate or relate to Solomonoff Induction.

We have outlined some relevant findings, showing that our hypotheses seem to be reflected in NNs in some ways but are complicated by computational limitations; we have then taken these findings into consideration to develop more complicated hypotheses that may paint a fuller picture of how Transformers approximate SolInd's optimal sequence prediction in practice.

We hope that future work in this area will help establish a common framework under which NNs can be considered and analysed.
% So yeah, this might be a helpful lens, let's see after we do some more work
% Well y'see, there's theoretical work we can do, and there's practical work

% We hope that this way of analysing and understanding Transformer models and sequence predictors in general can help establish a common methodology in research that seeks to understand how these systems work

\subsection{Future work}
There is much subsequent work to be done in this area towards investigating our hypotheses in more detail. The following may be useful next steps:
\begin{itemize}
    % \item Develop mathematical definitions of 
    \item Investigate whether Transformers can find shortcuts to, or better representations of, models of computation in between finite-state automata and TMs on the Chomsky hierarchy
    \item Determine whether sets of Transformers or other NNs of a given size that have been constructed to emulate TMs can be averaged to yield arbitrary weights, confirming that all NNs directly correspond to probability distributions over TMs
    \item Explore ways of decomposing Transformers into subnetworks that seem to emulate multiple functions using shared memory
    \item Convert decomposed subnetworks into explicit, non-NN-based algorithms
    \item Evaluate to what degree training an NN can be considered as part of the induction process, and to what degree it resembles selection of a UTM with short codes for patterns seen in the training corpus
\end{itemize}
% Develop, like, mathematical definitions of what it would mean for a TF to emulate SolInd

% Explore different ways of decomposing TFs into TMs, possible into classes that can be simulated together

%% file: samplepaper.bbl
\begin{thebibliography}{8}

\bibitem{tftm1}
Bhattamishra, S., Patel, A., Goyal, N.: On the computational power of transformers and its implications in sequence modeling. arXiv preprint, arXiv:2006.09286. (2020)

\bibitem{LLMsOnLLMs}
Bills, S., et al: Language models can explain neurons in language models. \url{https://openaipublic.blob.core.windows.net/neuron-explainer/paper/index.html}, last accessed 2024-07-22

\bibitem{chomsky}
Chomsky, N.: Three models for the description of language. IRE Transactions on information theory 2.3, pp. 113--124. (1956)

\bibitem{deletang}
Delétang, G., et al.: Neural networks and the Chomsky hierarchy. arXiv preprint, arXiv:2207.02098. (2022)

\bibitem{lottery}
Frankle, J., Carbin, M.: The lottery ticket hypothesis: Finding sparse, trainable neural networks. arXiv preprint, arXiv:1803.03635. (2018)

\bibitem{tfsolind}
Grau-Moya, J., et al.: Learning Universal Predictors. arXiv preprint, arXiv:2401.14953. (2024)

\bibitem{newbounds}
Hutter, M.: New error bounds for Solomonoff prediction. Journal of Computer and System Sciences 62.4, pp. 653--667. (2001)

\bibitem{shortcuts}
Liu, B., et al.: Transformers learn shortcuts to automata. arXiv preprint, arXiv:2210.10749. (2022)

\bibitem{decomp}
Liu, X., Parhi, K.: Tensor Decomposition for Model Reduction in Neural Networks: A Review [Feature]. IEEE Circuits and Systems Magazine, 23.2, pp.8--28. (2023)

\bibitem{tftm2}
Pérez, J., Marinković, J., Barceló, P.: On the Turing completeness of modern neural network architectures. arXiv preprint, arXiv:1901.03429. (2019)

\bibitem{shannon}
Shannon, C. E.: XXII. Programming a computer for playing chess. The London, Edinburgh, and Dublin Philosophical Magazine and Journal of Science, 41.314, pp. 256--275. (1950)

\bibitem{siegelmannsontag}
Siegelmann, H., Sontag, E.: On the computational power of neural nets Proceedings of the fifth annual workshop on Computational learning theory, pp. 440--449. (1992)

\bibitem{solind}
Solomonoff, R.: A formal theory of inductive inference. Part I. Information and control 7.1, pp. 1--22. (1964)

\bibitem{solind96}
Solomonoff, R.: Does algorithmic probability solve the problem of induction. Oxbridge Research, POB 391887. (1996)

\bibitem{goldengate}
Templeton, A., et al.: Scaling monosemanticity: Extracting interpretable features from Claude 3 Sonnet. \url{https://transformer-circuits.pub/2024/scaling-monosemanticity/index.html}, last accessed 2024-07-22

\bibitem{tf}
Vaswani, A., et al.: Attention is all you need. Advances in neural information processing systems 30.(2017)

\end{thebibliography}
